%% file: main.tex
\documentclass[final]{cvpr}

\usepackage{times}
\usepackage{epsfig}
\usepackage{graphicx}
\usepackage{amsmath}
\usepackage{amssymb}
\usepackage{booktabs}
\usepackage{mathtools}
\usepackage{multirow}
\usepackage{bbm}
\usepackage{xcolor}

\def\eg{\emph{e.g}.}
\def\ie{\emph{i.e}.}
\def\eg{\emph{e.g}.}
\def\ie{\emph{i.e}.}
\def\etal{\emph{et al.}}

\def\etc{\emph{etc}}

\newcommand{\RR}{\mathbb{R}}
\newcommand{\Exp}{\mathbb{E}}
\newcommand{\Dis}{\mathcal{D}}

\usepackage[pagebackref=true,breaklinks=true,colorlinks,bookmarks=false,urlcolor=black]{hyperref}

\def\cvprPaperID{3476} % *** Enter the CVPR Paper ID here
\def\confYear{CVPR 2021}
\input{math-com}

\begin{document}

%%%%%%%%% TITLE
\title{SelfDoc: Self-Supervised Document Representation Learning}

% \and

\author{Peizhao Li$^{1}$\thanks{This work was done during the author’s internship at Adobe Research.}, Jiuxiang Gu$^{2}$, Jason Kuen$^{2}$, Vlad I. Morariu$^{2}$, Handong Zhao$^{2}$, \\ Rajiv Jain$^{2}$, Varun Manjunatha$^{2}$, Hongfu Liu$^{1}$ \\
$^{1}$Brandeis University, $^{2}$Adobe Research \\
\small{\texttt{\{peizhaoli,hongfuliu\}@brandeis.edu}} \\
\small{\texttt{\{jigu,kuen,morariu,hazhao,rajijain,vmanjuna\}@adobe.com}}
}

\maketitle

%%%%%%%%% ABSTRACT
\begin{abstract}
We propose SelfDoc, a task-agnostic pre-training framework for document image understanding. Because documents are multimodal and are intended for sequential reading, our framework exploits the positional, textual, and visual information of every semantically meaningful component in a document, and it models the contextualization between each block of content. Unlike existing document pre-training models, our model is coarse-grained instead of treating individual words as input, therefore avoiding an overly fine-grained with excessive contextualization. Beyond that, we introduce cross-modal learning in the model pre-training phase to fully leverage multimodal information from unlabeled documents. For downstream usage, we propose a novel modality-adaptive attention mechanism for multimodal feature fusion by adaptively emphasizing language and vision signals. Our framework benefits from self-supervised pre-training on documents without requiring annotations by a feature masking training strategy. It achieves superior performance on multiple downstream tasks with significantly fewer document images used in the pre-training stage compared to previous works.
\end{abstract}

%%%%%%%%% BODY TEXT
\section{Introduction}

Documents, such as business forms, scholarly and news articles, invoices, letters, and text-based emails, encode and convey information through language, visual content, and layout structure. Automated document understanding is a crucial research area for business and academic values. It can significantly reduce labor-intensive document workflows through automated entity recognition, document classification, semantic extraction, document completion,~\etc.

Many works have been proposed applying machine learning for document analysis~\cite{jain2019multimodal,katti2018chargrid,yang2017learning,xu2020layoutlm,katti2018chargrid,liu2019graph}. However, parsing a document remains non-trivial and poses multiple challenges. One challenge is modeling and understanding contextual information when interpreting content. For example, since information in documents is organized for sequential reading, the interpretation of a piece of content relies heavily on its surrounding context. Similarly, a heading can indicate and summarize the meaning of subsequent blocks of text, and a caption could be useful for understanding a related figure. Another challenge is effectively incorporating the cues from multiple data modalities. In contrast to other data formats like images or plain text, documents combine textual and visual information, and both of the two modalities are complemented by the document layout. Additionally, from a practical perspective, many tasks related to document understanding are label-scarce. A framework that can learn from unlabeled documents (\textit{i.e.,} pre-training) and perform model fine-tuning for specific downstream applications is more preferred than the one that requires fully-annotated training data.

In this work, we develop a task-agnostic representation learning framework for document images. Our model fully exploits the textual, visual, and positional information of every semantically meaningful component in a document, \textit{e.g.}, text block, heading, and figure. To model the internal relationships among components in documents, we adopt the contextualized attention mechanism from natural language processing (NLP)~\cite{vaswani2017attention} and employ it at the component level. We design two branches separately for textual and visual representation learning, and later encourage cross-modal learning with the proposed cross-modality encoder. In order to seek a better modality fusion for downstream usage, we propose a modality-adaptive attention mechanism to fuse the language and vision features adaptively. Moreover, our framework learns a generic representation from a collection of unlabeled documents via self-supervised learning, and afterward, it will be fine-tuned on various document-related downstream applications.

There are two major differences between our SelfDoc and LayoutLM~\cite{xu2020layoutlm}, which also introduces a task-agnostic document pre-training framework by applying 2D positional encoding to BERT model~\cite{devlin2019bert}. 1) Instead of using \textit{word} as the basic unit for model input, we adopt semantically meaningful components (\textit{e.g.}, \textit{text block}, \textit{heading}, \textit{figure}) as the model input. In a document, a single word can be understood within the local context where it is found, and does not always require analyzing the entire page for every word. For instance, an answer in a questionnaire tends to be a complete sentence and already delivers semantics. Introducing the contextualization between every single word in documents may be redundant and also ignore localized context; 2) We advance the interaction between language and vision modality in the model's pre-training stage, therefore our model can efficiently leverage the multimodal information from unlabeled data. Comparatively, LayoutLM only considers a single modality in the pre-training stage and incorporates the visual clues during the fine-tuning phase.

We evaluate our model on three downstream tasks: document entity recognition, document classification, and document clustering. With the help of our pre-training method, we achieve leading performance on these applications over other pre-training and task-specific models. In short, our work contributes to the advancement of document analysis and intelligence by 1) introducing SelfDoc, a novel task-agnostic self-supervised learning framework for document data. Our model establishes the contextualization over a block of content and involves multimodal information; 2) modeling information from multiple modalities via cross-modal learning in the pre-training stage, and proposing a modality-adaptive attention mechanism to fuse language and vision features for downstream usages; 3) demonstrating superior performance by using fewer samples for pre-training. SelfDoc achieves surpassing performance on multiple downstream tasks comparing to other methods.

%-------------------------------------------------------------------------

\begin{figure*}
    \centering
    \includegraphics[width=1.\linewidth]{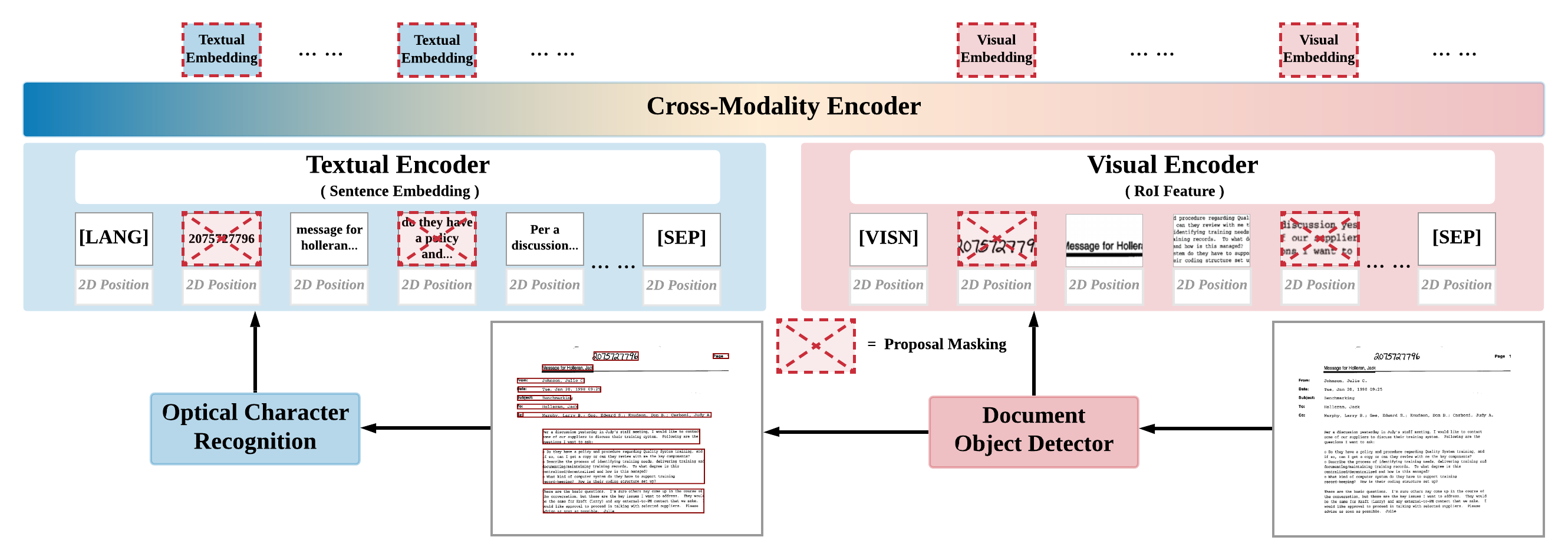}
    \vspace{-4.5mm}
    \caption{Overview of the proposed document representation learning framework. Extracted language and vision features with corresponding positional encoding are fed into a textual encoder, a visual encoder, and a cross-modal encoder to manipulate the contextual clues and multimodal information within documents. The produced features can be used for document analysis tasks.}
\label{fig:framework}
\vspace{-4.5mm}
\end{figure*}

%-------------------------------------------------------------------------

\section{Related Work}

\noindent\textbf{Document Image Understanding.} Artificial neural networks have been extensively applied to document analysis and recognition tasks like page segmentation, text location detection, and region labeling. Marinai~\etal~\cite{marinai2005artificial} survey connectionist-based approaches on document image processing. Moreover, Michael~\etal~\cite{shilman2005learning} propose a grammatical model for hierarchical document segmentation and labeling, while Hao~\etal~\cite{wei2013evaluation} utilize some statistical machine learning approaches to detect physical structures from historical documents. Recently, with deep learning showing great performance in many domains such as computer vision and NLP, some researchers have applied deep learning to document analysis~\cite{katti2018chargrid,liu2019graph,yang2017learning,jain2019multimodal,cosma2020self}. By utilizing the graphical property of documents, Katti \etal~\cite{katti2018chargrid} employ convolutional neural networks to recognize the bounding box and semantic segmentation in a document. Despite working with visual clues only, Yang~\etal~\cite{yang2017learning} use both convolutional neural networks and traditional textual embedding techniques to learn scanned documents by self-reconstruction. On handling the inner connections between text segments in invoices, Liu~\etal~\cite{liu2019graph} apply graph neural networks and manually build edges by similarity to model the inner-relations in documents. Jain and Wigington~\cite{jain2019multimodal} propose a multimodal ensemble approach combining language and vision models for document classification. These pioneering attempts toward applying machine learning to document data, though exciting and motivating, are heavily task-specific in their model design and require exhaustive annotations for document image representation learning.

\noindent\textbf{Self-Supervised Learning.} Pre-training models in NLP have shown great success in producing generic language representation that learns from a large scale of the unlabeled corpus. BERT~\cite{devlin2019bert}, which stands for bidirectional encoder representations from Transformers~\cite{vaswani2017attention}, and other pre-training models~\cite{peters2018deep,liu2019roberta,radford2018improving,gu2020self} have delivered promising performance on a series of downstream linguistic tasks such as question answering, sentence classification, and named entity recognition. The core idea of BERT is to learn a contextualized representation from corpus intrinsically via two self-supervised strategies. Given its success in NLP, some works extend the Transformer framework and model pre-training to vision-language learning~\cite{lu2019vilbert,tan2019lxmert}. These works focus on natural images and corresponding textual descriptions, learning the cross-modality alignment between visual and linguistic information, and can be applied to visual question answering~\cite{tan2019lxmert}, referring expression (localize an object with the given referring expression)~\cite{lu2019vilbert}, and image retrieval~\cite{gu2020self}. Although the cross-modality design in vision-language learning is used as an inspiration for document representation learning, it cannot be directly adapted for document data due to the great differences between document data and natural image data.

\noindent\textbf{Document Pre-training.} Most recently, some works have started pre-training models on document images~\cite{xu2020layoutlm,pramanik2020towards}. The first one, LayoutLM~\cite{xu2020layoutlm}, inherits the main idea from BERT while receiving the extra positional information for text in documents, and additionally includes image embeddings in the fine-tuning phase. Pramanik~\etal~\cite{pramanik2020towards} use Longformer~\cite{beltagy2020longformer} for heavily-word documents and extend the pre-training strategies to multi-page document pre-training. In~\cite{cosma2020self}, they introduce a document pre-training method by solving jigsaw puzzles and doing multimodal learning via topic modeling. Although they employ both image and text modalities during the training process, only image information is used when tested.
In contrast to these works, we establish our representation learning at the semantic-component level instead of the single word or character level in documents. By learning feature embedding on document components, we avoid the excessive contextualized learning between every word in a document but exploit the relations between each component. Beyond that, we introduce cross-modality learning in the pre-training phase for contextualized comprehension on document components across language and vision, and leverage multimodal information from document images without annotation.

%-------------------------------------------------------------------------

\section{Methodology}\label{sec:method}

Fig.~\ref{fig:framework} shows an overview of our SelfDoc representation learning framework. It takes document object proposals from a document object detector, and extracts features from both textual and visual modalities with positional encoding to serve as input. For each modality, we employ a single-modality encoder for contextual learning, and later perform learning over the two modalities using the proposed cross-modality encoder. The generated representation can be further utilized for downstream document understanding tasks, such as entity recognition or document classification.

%-------------------------------------------------------------------------

\subsection{Pre-processing and Feature Extraction}

To begin with, we train a document object detector using Faster R-CNN~\cite{ren2015faster} on public document datasets\cite{zhong2019publaynet,mathew2020docvqa} with bounding box annotations on semantically meaningful components, and localize significant components (\ie, document object proposals) of a document. In our current implementation, we detect the following categories: \textit{text block}, \textit{title}, \textit{list}, \textit{table}, and \textit{figure}. We deem the detected proposals as the basic input unit of our framework. Next, we apply an Optical Character Recognition (OCR) engine~\cite{10.5555/1288165.1288167} to process each cropped proposal from the original document and get the detected text in a default word order.

We then extract the textual and visual features for each proposal. For textual features, we embed plain text contained in a proposal into a feature vector using the pre-trained Sentence-BERT model~\cite{reimers2019sentence}: a sentence and sequential word learning model that demonstrates superior performance on semantic textual similarity and sentence classification tasks. We extract visual features from Regions-of-Interest (RoI) heads in Faster R-CNN model for every detected proposal. The RoI head uses an adaptive pooling function to output a fixed size vector for proposals of arbitrary sizes. Formally, a document $D = \{p_1, \ldots, p_N\}$ consists of $N$ document object proposals, where each object proposal $p_i = \{x_\text{pos}^i\in\RR^4, x_\text{visn}^i\in\RR^{d_\text{visn}}, x_\text{lang}^i\in\RR^{d_\text{lang}}\}$ is represented by its 2D coordinate $x_\text{pos}$, its RoI feature $x_\text{visn}$, and sentence embedding for text $x_\text{lang}$, with corresponding feature dimensions $d_\text{lang}$ and $d_\text{visn}$ respectively.

Compared to word-level input, the component-level formulation can reduce the input sequence length for a document, especially for text-heavy documents such as scholarly articles. Therefore it decreases the amount of time needed for training and inference since the time complexity for a fully contextualized attention operation (that will be described later) scales quadratically with the input length.

% After employing OCR, sequential words \jx{$S_{p_n} = \{w_1, \ldots, w_{N_{p_n}}\}$} for a proposal is given with a default order by OCR (usually from left to right and top to bottom), \jx{where $N_{p_n}$ is the number of words for each proposal}.
%We generate embedding for a proposal $p_n = \{x^\text{lang}, x^\text{visn}, pos\}$ by $x^\text{lang} = \text{Sentence-Bert}(W_{p_n})$ and $x^\text{visn}$, $pos$ given by detector.

%-------------------------------------------------------------------------

\subsection{Input Modeling}\label{sec:input}

Inspired by BERT~\cite{devlin2019bert}, we mark the beginning of a sentence sequence with a special [LANG] token and an RoI region sequence with [VISN] token (shown in Fig.~\ref{fig:framework}), which are respectively calculated by averaging the sentence features and RoI features. Also, we manually set the positional coordinate of these special proposals to cover the whole document page. The input sequence is then zero-padded to match with its batch-peer for batch training. Then, to incorporate positional information, the input features are mapped to hidden states $\mH^0_T = \{h_T^1, \ldots, h_T^N\}$ and $\mH^0_V = \{h_V^1, \ldots, h_V^N\}$ by a linear mapping as follows:
\begin{equation}
    h_T^i = \mW_T x_\text{lang}^i + \mW_P x_\text{pos}^i,\ 
    h_V^i = \mW_V x_\text{visn}^i + \mW_P x_\text{pos}^i,
\end{equation}
where matrices $\mW_T\in\RR^{d_h\times d_\text{lang}}$, $\mW_V\in\RR^{d_h\times d_\text{visn}}$, $\mW_P\in\RR^{d_h\times 4}$ project features into hidden-state in $d_h$ dimension.

%-------------------------------------------------------------------------

\begin{figure*}
\small
    \centering
    \includegraphics[width=1.\linewidth]{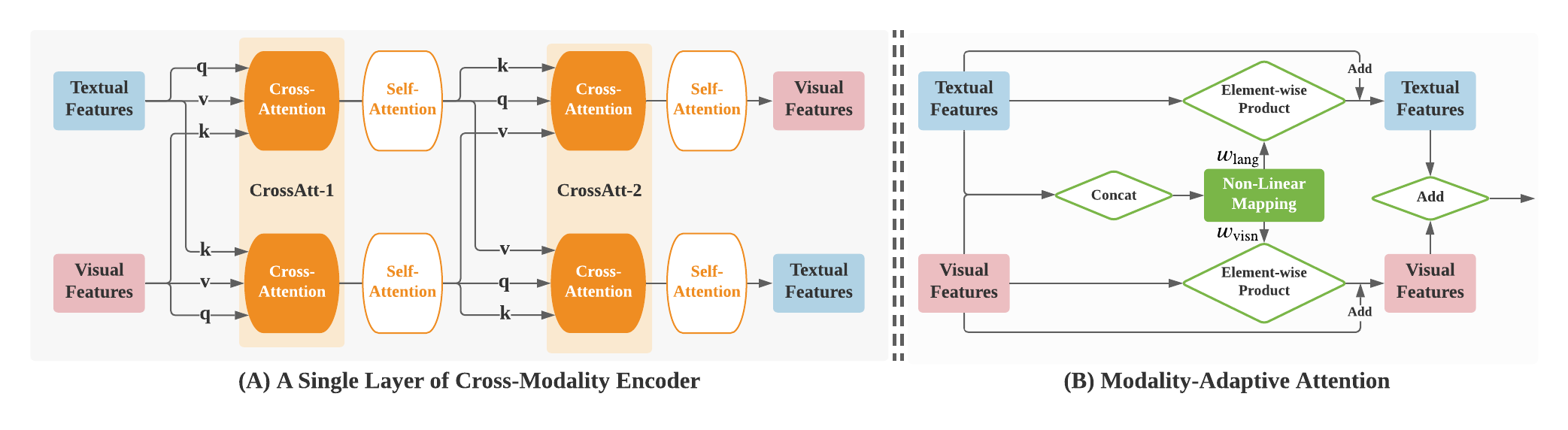}
    \vspace{-4.5mm}
    \caption{Schematic illustrations of (A): cross-modality encoder with different cross-attention functions inserted sequentially, and (B): modality-adaptive attention used in the fine-tuning phase for adaptively fusing features from language and vision.}
    \vspace{-4.5mm}
\label{fig:module}
\end{figure*}

%-------------------------------------------------------------------------

\subsection{Single-Modality Encoder}

Then, the language and vision features are separately passed through the textual and visual encoder. These two encoders follow the same design of a basic module in BERT, but their parameters are not shared. This module contains multi-head attention, feed-forward (FF) layers, residual connection, and layer normalization (LN)~\cite{ba2016layer}. In the multi-layer single-modality encoder, let $\mH^l = \{h_1, \ldots, h_N\}$ be the encoded features at the $l$-th layer. $\mH^0$ is the vector of the input features given in Sec.~\ref{sec:input}. Features output by the next layer $\mH^{l+1}$ can be obtained via:
\begin{align}\label{eq:self}
    &\mH^{l}_\text{att} = \text{LN}(f_\text{SelfAtt}(\mH^l) + \mH^l), \\
    &\mH^{l+1} = \text{LN}(f_\text{FF}(\mH^{l}_\text{att}) + \mH^{l}_\text{att}),
\end{align}
where $f_\text{SelfAtt}(\cdot)$ is the self-attention function defined as
\begin{equation}
    f_\text{SelfAtt}(\mH^l) = \text{softmax}\left({\frac{q(\mH^l)k(\mH^l)^\top}{\sqrt{d_k}}}\right)v(\mH^l),
    % \text{Attention}(Q,K,V) \coloneqq \text{softmax}(\frac{Q K^\top}{\sqrt{d_k}})V,
\end{equation}
where $q(\cdot)$, $k(\cdot)$, and $v(\cdot)$ are linear transformation layers applied to features of proposals and they are the query, key, and value, respectively. $d_k$ is the number of attention heads for normalization. For technical details on multi-head attention please refer to~\cite{vaswani2017attention}. Finally, $\mH^{l+1}$ can be obtained by $\mH^{l}_\text{att}$ via a feed-forward sub-layer composed of two fully-connected layers of function $f_\text{FF}(\cdot)$. Hierarchically stacked layers form the textual and visual encoders.

% is , LN is layer normalization. Self-attention function $f_\text{SelfAtt}(H^l)$ is defined as
% \begin{equation}
%     f_\text{SelfAtt}(\mH^l) = \text{softmax}({\frac{q(\mH^l)k(\mH^l)}{\sqrt{d_k}})}v(\mH^l)
%     % \text{Attention}(Q,K,V) \coloneqq \text{softmax}(\frac{Q K^\top}{\sqrt{d_k}})V,
% \end{equation}
% where $q(\cdot)$, $k(\cdot)$, and $v(\cdot)$ is linear layer transformation referred as query, key, and value. $d_k$ is the number of attention head for normalization. Hierarchically stack multiple basic modules form the textual or visual encoder.

The textual and visual encoders produce contextually embedded features for each proposal using the features from surrounding proposals in their respective modality. The vector multiplication between query and key explores similar patterns in sequential proposals and emphasizes the shared part. The outputs of the textual and visual encoder are subsequently fed into the cross-modal encoder described below, which is more focused on cross-modality learning to bridge the multimodal information in language and vision.

%-------------------------------------------------------------------------

\subsection{Cross-Modality Encoder}

We encourage cross-modality learning by introducing two interactive cross-attention functions. The structure of the cross-modal encoder is similar to the textual or visual encoder, but we substitute the self-attention function $f_\text{SelfAtt}(\cdot)$ in Eq.~(\ref{eq:self}) with $f_\text{CrossAtt-1}(\cdot)$ or $f_\text{CrossAtt-2}(\cdot)$ as elaborated in what follows. We add subscripts $T$ and $V$ to denote the modality in $\mH^l_{\text{T}}$ and $\mH^l_{\text{V}}$ which are the intermediate textual and visual representations, respectively.

The first attention function identifies the agreement between language and vision information. At a high level, if the font size or character style in a proposal is confirmed by the semantic meaning of language features, these features should be amplified. Formally, we have 
\begin{align}
    %f_\text{CrossAtt}(\mH^l_\text{lang}) &= \text{softmax}({\frac{q(\mH^l_\text{lang})k(\mH^l_\text{visn})}{\sqrt{d_k}})}v(\mH^l_\text{lang}) \\
    f_\text{CrossAtt-1}(\mH^l_\text{T}) &= \text{softmax}\left({\frac{q(\mH^l_\text{T})k(\mH^l_\text{V})^\top}{\sqrt{d_k}}}\right)v(\mH^l_\text{T}), \\
    %f_\text{CrossAtt}(\mH^l_\text{visn}) &= \text{softmax}({\frac{q(\mH^l_\text{visn})k(\mH^l_\text{lang})}{\sqrt{d_k}})}v(\mH^l_\text{visn})
    f_\text{CrossAtt-1}(\mH^l_\text{V}) &= \text{softmax}\left({\frac{q(\mH^l_\text{V})k(\mH^l_\text{T})^\top}{\sqrt{d_k}}}\right)v(\mH^l_\text{V}).
\end{align}

The second attention function serves as the operation to discover inner-relationships from one modality to another. Since documents are naturally composed of two modalities, we have the same number of proposals in language and vision branches, and the input sequences of these two modalities are identically ordered\footnote{Identically ordered means the two modalities have the same input order, \eg, the first index of textual and visual input corresponds to the same proposal.} in our input. Based on this, the contextual clues can be propagated between modalities, for instance, the similarity in font style can enhance the understanding of semantic meaning between proposals. We have
\begin{align}
    % f_\text{CrossAtt}(H^l_\text{lang}) &\coloneqq \text{softmax}({\frac{q(H^l_\text{visn})k(H^l_\text{visn})}{\sqrt{d_k}})}v(H^l_\text{lang}) \\
    f_\text{CrossAtt-2}(\mH^l_\text{T}) &= \text{softmax}\left({\frac{q(\mH^l_\text{V})k(\mH^l_\text{V})^\top}{\sqrt{d_k}}}\right)v(\mH^l_\text{T}), \\
    %f_\text{CrossAtt}(H^l_\text{visn}) &\coloneqq \text{softmax}({\frac{q(H^l_\text{lang})k(H^l_\text{lang})}{\sqrt{d_k}})}v(H^l_\text{visn})
    f_\text{CrossAtt-2}(\mH^l_\text{V}) &= \text{softmax}\left({\frac{q(\mH^l_\text{T})k(\mH^l_\text{T})^\top}{\sqrt{d_k}}}\right)v(\mH^l_\text{V}).
\end{align}

Note that we do not distinguish the linear transformation query, key, and value for notational simplicity, but they are not shared and are specific for a certain attention function in implementation. We build the cross-modality encoder by alternatively inserting these two types of cross-attention layer and a self-attention layer. A schematic illustration for a cross-modality encoder is presented in Fig.~\ref{fig:module}(A).

%-------------------------------------------------------------------------

\subsection{Pre-training}

Our learning framework can benefit from documents without annotations via a self-supervised training strategy. In the pre-training stage, we have a masking function $f_\text{Mask}(\cdot)$ that randomly mask selected proposals in a document in language or vision branch with a pre-defined probability that can 1) set the language or vision feature to zeros, or 2) replace the feature with a random proposal in the same modality from the pre-training corpus, or 3) keep the original feature unchanged. In the pre-training stage, we minimize the pre-training objective function as follows:
\begin{equation}
\begin{split}
    L = & \Exp_{\Dis_\text{Mask}^\text{lang}} L_1(x_\text{lang}^i - f_\text{SelfDoc}(f_{\text{Mask}}(x_\text{lang}^i) | D)) \\ 
    + & \Exp_{\Dis_\text{Mask}^\text{visn}} L_1(x_\text{visn}^i - f_\text{SelfDoc}(f_{\text{Mask}}(x_\text{visn}^i) | D)),
\end{split}
\end{equation}
where $\Dis_{\text{Mask}}^\text{lang}$ and $\Dis_{\text{Mask}}^\text{visn}$ are the distributions for masked proposals on language and vision branches, respectively. $f_\text{SelfDoc}(\cdot)$ denotes our whole model which outputs a feature embedding for each proposal. $L_1$ represents Smooth L1 loss~\cite{girshick2015fast}. $D$ represents the document and, in this context, can be viewed as the surrounding features of proposals of the masked features from the two modalities.

The proposal selection and masking function are applied independently to language and vision features. The pre-training objective function working with modality interaction not only infers the masked features from surrounding proposals in the same modality, but can also absorb features from another modality and encourage cross-modal learning.

%-------------------------------------------------------------------------

\subsection{Modality-Adaptive Attention}

In the fine-tuning phase and downstream usage, we fuse the output features for each proposal from both language and vision modalities. Most previous multimodal works~\cite{jain2019multimodal,xu2020layoutlm} use a simple linear additive operation for fusion. Considering the diverse variety of document images, we propose a modality-adaptive attention (M-AA) for a better feature fusion. The general idea is to apply sample-dependent attention weights to the two modalities and emphasize or diminish the intensity of language or vision features adaptively for different documents. Intuitively, this input-dependent attention can be helpful on some samples in raw documents such as: 1) ones that contain handwriting that is not recognizable by OCR algorithms, in which case a stronger emphasis on visual clue is needed; 2) documents that already contain abundant linguistic information such as scholarly articles, in which case a stronger emphasis on the semantic meaning of language is more helpful.

We summarize the pipeline for this module in Fig.~\ref{fig:module}(B). To be specific, we concatenate the output features of each proposal from language and vision branches, and feed it to a non-linear mapping network $\RR^{2\times d_h}\rightarrow \RR^{2}$ ($d_h$ is the dimension of output features in either language or vision branch), then split the output weights into $w_\text{lang}\in\RR^{1}$ and $w_\text{visn}\in\RR^{1}$, and return the weights separately to its respective modality to perform element-wise product. We multiply the language and vision features with their modality-specific attention weight, then after a residual connection, features from two modalities are fused by a linear additive function. In our implementation, we employ a two-layer neural network that ends with a sigmoid activation function to achieve non-linear mapping.

%-------------------------------------------------------------------------

\section{Experiments}\label{sec:exp}

%-------------------------------------------------------------------------

\begin{table*}
\small
    %\caption{Experimental results and comparison on document entity recognition in FUNSD dataset and document classification in RVL-CDIP dataset. `M-AA' denotes the modality-adaptive attention module. The symbol $\ddag$ implies feature fusing with global visual features on the whole document images from VGG-16. We re-implement LayoutLM on document classification and denote the result as `our impl', while the better results denote as $\sharp$ is achieved using another data source. Please refer to the paragraph document classification in Sec.~\ref{sec:app} for explanation on `our impl.' and $\sharp$.}
    \caption{Experimental results and comparison on document entity recognition in FUNSD dataset and document classification in RVL-CDIP dataset. The symbol $\ddag$ implies feature fusing with global visual features on the whole document images from VGG-16. We re-implement LayoutLM on document classification and denote the result as `our impl.', while the better results denote as $\sharp$ is achieved using another data source. Please refer to the paragraph document classification in Sec.~\ref{sec:app} for explanation on `our impl.' and $\sharp$.}
    \centering
    \small
    \setlength{\tabcolsep}{3.5pt}{
    \begin{tabular}{l|ccc|c|c}
        \toprule
        
        Method & \# Pre-training Data & Modality & Architecture & Entity Recognition & Classification \\
        
        \midrule
        
        $\text{VGG-16}$ & - & Vision & - & - & 0.9031 \\
        $\text{ResNet-50}$ & - & Vision & - & - & 0.8866 \\
        $\text{Multimodal Ensemble}$~\cite{dauphinee2019modular} & - & Language + Vision & MLP + VGG-16 & - & 0.9303 \\
        $\text{Jain and Wigington}$~\cite{jain2019multimodal} & - & Language + Vision & MLP + VGG-16 & - & 0.9360 \\
        $\text{Sentence-BERT}$~\cite{reimers2019sentence} & - & Language & - & 0.6947 & - \\
        
        \midrule
        
        $\text{BERT}_{\text{BASE}}$~\cite{devlin2019bert} & - & Language & - & 0.6062 & 0.8610 \\
        $\text{RoBERTa}_{\text{BASE}}$~\cite{liu2019roberta} & - & Language & - & 0.6648 & 0.8682 \\
        $\text{Pramanik}$~\etal~\cite{pramanik2020towards} & 110K & Language + Vision + Layout & - & 0.7744 & 0.9172 \\
        $\text{LayoutLM}_{\text{BASE}}$~\cite{xu2020layoutlm} & 500K & Language + Layout & - & 0.6985 & 0.9125 \\
        $\text{LayoutLM}_{\text{BASE}}$~\cite{xu2020layoutlm} & 1M & Language + Layout & - & 0.7299 & 0.9148 \\
        $\text{LayoutLM}_{\text{BASE}}$~\cite{xu2020layoutlm} & 2M & Language + Layout & - & 0.7592 & 0.9165 \\
        $\text{LayoutLM}_{\text{BASE}}$~\cite{xu2020layoutlm} & 11M & Language + Layout & - & 0.7866 & 0.9178 \\
        $\text{LayoutLM}_{\text{LARGE}}$~\cite{xu2020layoutlm} & 1M & Language + Layout & - & 0.7585 & 0.9188 \\
        $\text{LayoutLM}_{\text{LARGE}}$~\cite{xu2020layoutlm} & 11M & Language + Layout & - & 0.7789 & 0.9190 \\
        $\text{LayoutLM}_{\text{BASE}}$~\cite{xu2020layoutlm} & 1M & Language + Vision + Layout & - & 0.7441 & $\text{0.9431}^\sharp$ \\
        $\text{LayoutLM}_{\text{BASE}}$~\cite{xu2020layoutlm} & 11M & Language + Vision + Layout & - & 0.7927 & $\text{0.9442}^\sharp$ \\
        
        \midrule
        
        $\text{LayoutLM}_{\text{BASE}}$ (our impl.) & 11M & Language + Layout & - & 0.7887 & 0.8857 \\
        $\text{LayoutLM}_{\text{BASE}}$ (our impl.) & 11M & Language + Vision + Layout & - & 0.7993 & $\text{0.9169}^\ddag$ \\
        
        \midrule
        
        $\text{SelfDoc}$ & Scratch & Language + Vision + Layout & w/o M-AA & 0.7607 & 0.9049 \\
        $\text{SelfDoc}$ & 320K & Language + Vision + Layout & w/o M-AA & 0.8263 & 0.9263/$0.9364^\ddag$ \\
        $\text{SelfDoc}$ & 320K & Language + Vision + Layout & with M-AA & \textbf{0.8336} & 0.9281/$\textbf{0.9381}^\ddag$ \\
        
        \bottomrule
    \end{tabular}}
    \label{tab:res1}
    \vspace{-2mm}
\end{table*}

%-------------------------------------------------------------------------

\subsection{Implementation}\label{sec:impl}

\begin{figure}
    \centering
    \includegraphics[width=1.\columnwidth]{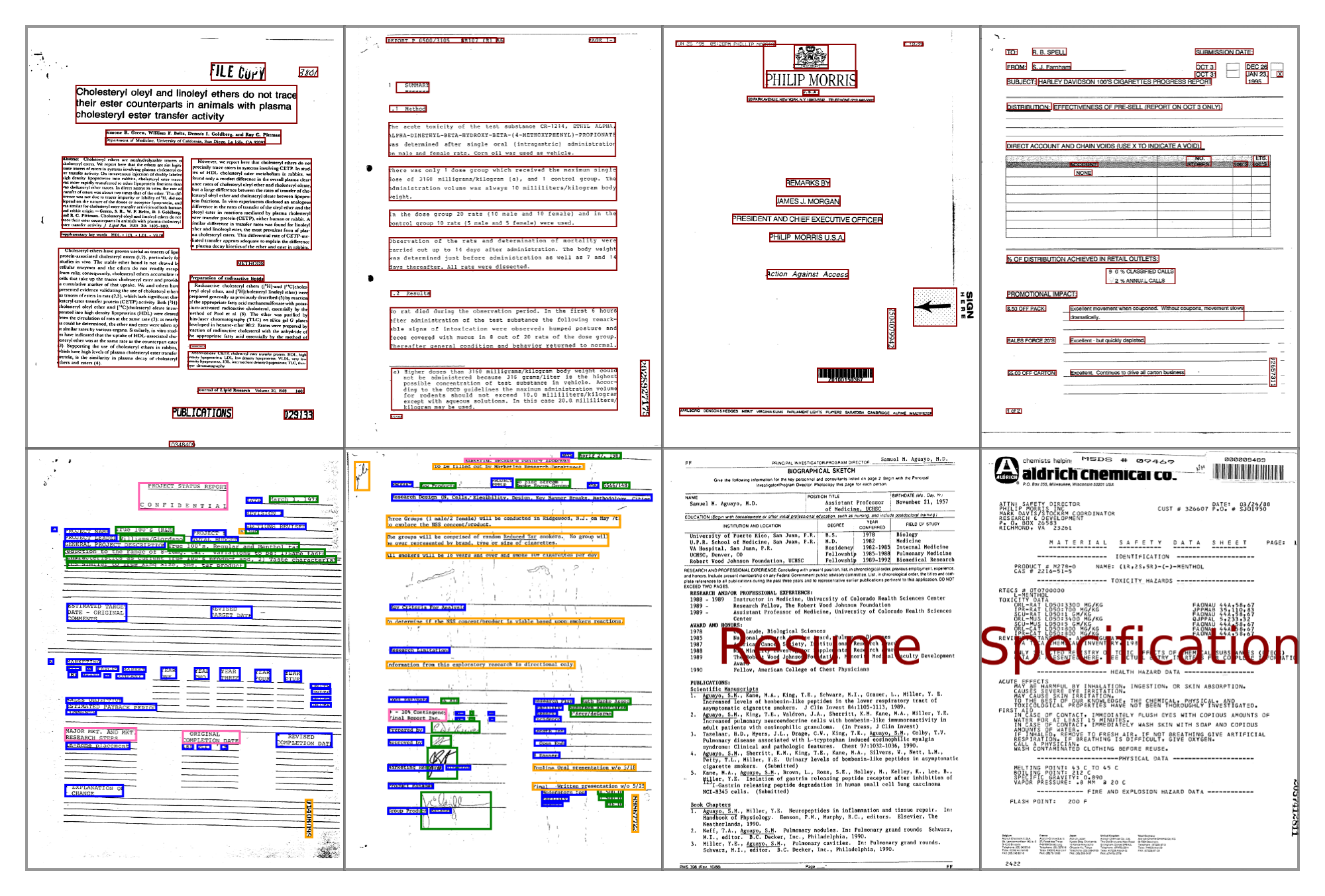}
    \vspace{-4mm}
    \caption{Example outputs. First row: examples of document object detection. Second row: examples of document entity recognition with different colors indicating different entity categories (the left two), and document classification with labels (the right two).}
    \label{fig:showcase}
    \vspace{-4mm}
\end{figure}

\noindent\textbf{Dataset}. We use the PubLayNet dataset~\cite{zhong2019publaynet} and DocVQA dataset~\cite{mathew2020docvqa} to train the document object detector. PubLayNet includes 340K scholarly articles with bounding box on \textit{text block}, \textit{heading}, \textit{figure}, \textit{list}, and \textit{table}, and DocVQA has 12K forms with a bounding box annotated for each text block. We use the official OCR results provided by the DocVQA website as the bounding boxes for text blocks to train the detector. We pre-train SelfDoc on the RVL-CDIP dataset~\cite{harley2015icdar}, a document classification dataset containing 320K documents for training, 40K for validation, and 40K for testing. Pre-training is only conducted on the training set. Fig.~\ref{fig:showcase} shows some image samples in document object detection, entity recognition, and document classification.

\noindent\textbf{Document pre-processing}. We train the document object detector using Detectron2~\cite{wu2019detectron2} with the ResNeXt-101~\cite{xie2017aggregated} backbone model. We apply rotations on images as data augmentation to improve the detection of the potential vertical text in documents. After obtaining the detection results, we use Tesseract OCR~\cite{10.5555/1288165.1288167}, a public OCR engine, to extract the plain text from each proposal given by detector. We crop the proposals from the original document images, and expand the bounding box by a factor of 1.1 and apply $2\times$ image magnification to better recognize the characters close to the edge and the overall word recognition. We convert detected OCR results into lower case, and convert digits to words. Common contractions are expanded before tokenization. For sentence embedding, we use the pre-trained sentence encoder (\texttt{bert-large-nli-mean-tokens})\footnote{https://github.com/UKPLab/sentence-transformers}. For visual feature extraction, we concatenate the feature from the last and P2 layers (second to last convolutional layer) in RoI heads. We have $d_\text{visn}=\text{2048}$ and $d_\text{lang}=\text{1024}$.

\noindent\textbf{Pre-training}. Due to their variety, the number of proposals in documents may vary significantly. This variance could cause some input sequences to be heavily padded to ensure that all sequences are as long as the longest sequence in batch-wise training, therefore slowing down the training speed. To deal with this issue, we do not only set a maximum length of input sequences, but also apply batch thresholding to avoid excessive padding for some documents and reduce some batches in the sequence length when feasible. Every batch contains documents that have the number of proposals concurrently below or beyond the threshold. We set the maximum length of proposals to 50, and the group threshold to 30. When proposals exceed the maximum limitation, we randomly sample the proposals in this document. The input sentence sequence and RoI sequence are ended with the special token [SEP], respectively. For the model architecture, we assign 4 layers to each of the textual and visual encoders, and continue with 2 cross-modality layers. The total number of layers for each branch is 12 and is equivalent to $\text{BERT}_{\text{BASE}}$. We keep other specific architectural configurations and masking probability the same as in $\text{BERT}_{\text{BASE}}$. In the pre-training phase, we set the batch size to 768, learning rate to 1e-4 in AdamW optimizer~\cite{loshchilov2018decoupled} with a linear warm-up ratio to 0.05 and linear decay. We do not initialize our model before pre-training with parameters from pre-trained BERT or any variants. We conduct pre-training for 72K iterations on 8 Tesla V-100 GPUs and it takes around 21 hours to complete the pre-training.

%-------------------------------------------------------------------------

%-------------------------------------------------------------------------

\begin{table*}
    \caption{Experimental results on document clustering over different number of cluster centroids and samples.}
    \centering
    \small
    \setlength{\tabcolsep}{7.5pt}{
    \begin{tabular}{l|cc|cc|cc|cc|cc}
        \toprule
        
        \multirow{2}{*}{Method} & \multicolumn{2}{|c|}{\# Cluster = 4} & \multicolumn{2}{|c|}{\# Cluster = 6} & \multicolumn{2}{|c|}{\# Cluster = 8} & \multicolumn{2}{|c|}{\# Cluster = 10} & \multicolumn{2}{|c}{\# Cluster = 12} \\
        & Acc. & NMI & Acc. & NMI & Acc. & NMI & Acc. & NMI & Acc. & NMI \\
        
        \midrule
        
        $\text{BERT}_{\text{BASE}}$~\cite{devlin2019bert} & 0.4190 & 0.1589 & 0.3743 & 0.1568 & 0.3400 & 0.1670 & 0.3903 & 0.2752 & 0.3021 & 0.2378 \\
        $\text{LayoutLM}_{\text{BASE}}$~\cite{xu2020layoutlm} & 0.4380 & 0.2107 & 0.4147 & 0.2111 & 0.3612 & 0.1821 & 0.3237 & 0.2213 & 0.2646 & 0.1979 \\
        
        \midrule
        
        Input Embedding & 0.4700 & 0.1946 & 0.3937 & 0.1723 & 0.3898 & 0.2636 & 0.3457 & 0.2549 & 0.3033 & 0.2315 \\
        SelfDoc & \textbf{0.6970} & \textbf{0.4155} & \textbf{0.4550} & \textbf{0.3025} & \textbf{0.4476} & \textbf{0.3473} & \textbf{0.4010} & \textbf{0.3612} & \textbf{0.3614} & \textbf{0.3173} \\
        
        \bottomrule
    \end{tabular}}
    \label{tab:res2}
    \vspace{-2mm}
\end{table*}

%-------------------------------------------------------------------------

\subsection{Applications}\label{sec:app}

\noindent\textbf{Document entity recognition.} The first downstream task for evaluation is document entity recognition. We adopt the FUNSD dataset~\cite{8892998} instead of SROIE~\cite{sroie} since FUNSD is an in-domain dataset. It contains 149 forms with 7,441 entities for fine-tuning, and 50 forms with 2,332 entities for testing. Semantic entities are classified into four categories: `\textit{header}', `\textit{question}', `\textit{answer}', or `\textit{other}'. The bounding box and plain text are given for each block of text as well for every single word, but we only make use of the positional and textual information for text blocks. We extract visual features from the RoI head in detector using the given box coordinates for each block, and extract the language embedding using the plain text as described previously. Micro averaged F1 score is used as the evaluative metric. We train a linear classifier on the output of our pre-trained model in the fine-tuning phase, with parameters in the pre-trained model fixed. The fine-tuning phase takes 60 epochs with a learning rate of 5e-5 and batch size of 16. Note that the experiment of FUNSD in LayoutLM~\cite{xu2020layoutlm} considers the prediction of word positions (begin, intermediate, end). Since our model considers the whole block of text, the word position is obvious and results remain the same in that setting.

\noindent\textbf{Document classification.} We use the RVL-CDIP~\cite{harley2015icdar} dataset to evaluate the performance of document classification, using the training set for fine-tuning and the testing set for evaluation. It consists of 400k images in 16 classes. We take the encoder outputs on the special tokens [LANG] and [VISN] from the modality-adaptive attention module as holistic representations of the textual and visual inputs. The addition of two features is used as the input to the classifier. The whole fine-tuning takes 20K iterations with a batch size of 768 and a learning rate of 5e-5.

Several technical issues related to the baseline model LayoutLM~\cite{xu2020layoutlm} in this task, drive us to re-implement this model to make a fair comparison. 1) In their experiments, LayoutLM does not use document images from RVL-CDIP, instead, they retrieve the corresponding images from IIT-CDIP test collection 1.0~\cite{10.1145/1148170.1148307}, which is a superset of RVL-CDIP but contains high-quality document images. The most obvious advantage is on the detected results of OCR, where the text is cleaner and more informative. Unfortunately, we had issues~\cite{layoutlmgithub2020} accessing IIT-CDIP; 2) LayoutLM~\cite{xu2020layoutlm} uses an image embedding from a detector over the whole document image on this task, and jointly trains the detector during fine-tuning. However, the released code does not contain the jointly fine-tuned detector model. Given these two technical issues, we built our implementation using the released pre-trained models $\text{LayoutLM}_\text{BASE}$ and fine-tuning protocols to make the result a fair comparison, and denote it as `our impl.'. We also tried to fine-tune the released $\text{LayoutLM}_\text{LARGE}$ model but faced a convergence issue due to the need to restrict the batch size for computational reasons.

%\footnote{https://github.com/microsoft/unilm/issues/247}.
%\footnote{https://github.com/microsoft/unilm/issues/250}.

We hereby provide our model and LayoutLM with the same source of data, and also provide the results fusing with the same image embedding from VGG-16~\cite{simonyan2014very} that trained on RVL-CDIP. We choose the embedding from VGG-16 instead of other sophisticated models since some previous work~\cite{dauphinee2019modular,jain2019multimodal} use VGG-16 on document classification.

%-------------------------------------------------------------------------

\begin{table*}    
    \small
    \caption{Ablation studies on SelfDoc on multimodal information, pre-training data, and cross-modality attention.}
    \centering
    \setlength{\tabcolsep}{12pt}{
    \begin{tabular}{l|c|c|c|c}
        \toprule
        
        Setting & Modality & Parameter & Entity Recognition & Classification \\
        
        \midrule
        
        Scratch, remove layout  & Language + Vision & 137M & 0.7579 & 0.9070 \\
        Scratch, remove language  & Vision + Layout & 60M & 0.6209 & 0.8447 \\
        Scratch, remove vision  & Language + Layout & 60M & 0.7491 & 0.8895 \\
        Pre-train with 40K data  & Language + Vision + Layout & 137M & 0.7886 & 0.9119 \\
        Pre-trained w/o CrossAtt-1\&2 & Language + Vision + Layout & 146M & 0.7911 & 0.9189 \\
        Pre-trained w/o CrossAtt-1  & Language + Vision + Layout & 137M & 0.8152 & 0.9224 \\
        Pre-trained w/o CrossAtt-2  & Language + Vision + Layout & 137M & 0.8130 & 0.9229 \\
        
        \bottomrule
    \end{tabular}}
    \label{tab:abl}
    \vspace{-2mm}
\end{table*}

%-------------------------------------------------------------------------

\noindent\textbf{Document clustering.} We also investigate models in the scenario where there is no annotation available. We consider document clustering on RVL-CDIP~\cite{harley2015icdar} testing set. We randomly sample from the testing set and create five experimental scenarios, with \{3, 5, 7, 10, 12\} clusters. The corresponding numbers of samples are \{1k, 3k, 5k, 7k, 9k\}. In this task, we also include the input embedding (sentence embedding and visual features with the layout) of our model for comparison. Since all models do not have a supervision or pre-training objective function at the document image level, we take the average of input proposal sequence or word sequence as a representation of the document, and conduct K-means~\cite{arthur2006k} clustering over all the document representations. The model for pre-trained $\text{LayoutLM}_\text{BASE}$ is directly used. We use metrics clustering accuracy (Acc.) and normalized mutual information (NMI) for evaluation.

%-------------------------------------------------------------------------

\subsection{Result and Discussion}\label{sec:res}

Quantitative results are listed in Table~\ref{tab:res1} \&~\ref{tab:res2}, and an ablation study on SelfDoc is presented in Table~\ref{tab:abl}. We discuss our observations from the experiments as follows.

\noindent\textbf{Baselines.} We include five task-specific baselines in Table~\ref{tab:res1}. These include two standard convolutional neural networks VGG-16 and ResNet-50, two multimodal ensemble approaches~\cite{jain2019multimodal,dauphinee2019modular} using VGG-16 and a neural network for text encoding, plus the Sentence-BERT embedding. We have four task-agnostic learning methods, including two pre-trained language models~\cite{devlin2019bert,liu2019roberta}, the approach proposed by Pramanik~\etal~\cite{pramanik2020towards} pre-trained on arXiv dataset~\cite{arxiv}, and LayoutLM~\cite{xu2020layoutlm} pre-trained on IIT-CDIP dataset~\cite{10.1145/1148170.1148307}.

\noindent\textbf{SelfDoc outperforms baselines.} Table~\ref{tab:res1}~\&~\ref{tab:res2} show that SelfDoc outperforms baselines on document entity recognition, document classification and clustering. The only result we cannot outperform is the reported number by LayoutLM which uses a cleaner data source and jointly fine-tunes with a deeper CNN model, as we discussed in Sec.~\ref{sec:app}. Other than that, SelfDoc demonstrates a good performance on all three evaluative downstream tasks with significantly fewer data for pre-training. We deem the effective usage of pre-training documents as a superior advantage of our model since documents often contain sensitive information, thus the legal and privacy issues may limit the feasibility to build large scale high-quality document datasets for pre-training.

\noindent\textbf{A proposal is richer than a single word.} From the document clustering results in Table~\ref{tab:res2}, we observe that our input embedding is able to deliver a more informative representation than BERT and LayoutLM. This indicates that our pre-trained model can further improve the discriminability in features without fine-tuning. Experiments on clustering suggest that on the representation of the whole document image, exploring information from proposals can be more helpful than collecting features from each word. In addition, well-designed modeling on feature embedding can also bring informative representation.

%-------------------------------------------------------------------------

\noindent\textbf{Multimodal modeling is beneficial.} In the first three rows of our ablation study in Table~\ref{tab:abl}, we consider the scenario where removal happens on the three input components of SelfDoc: textual input, visual input, and structural layout, one at a time. We observe a significant drop in performance when removing textual or visual input on both entity recognition and classification, confirming the necessity for learning on two modalities. The structural layout has a smaller effect but it still contributes to the classification.

\noindent\textbf{Effectiveness of cross-modal learning.} In the last three rows of Table~\ref{tab:abl}, we investigate the effectiveness of cross-modal learning in our model. Note that we do not shrink the number of model parameters when removing a cross-modality attention function. When removing a single attention function (denoted as w/o CrossAtt-1/2), we fill the space with the remaining attention function. When totally removing the cross-attention encoder, we deepen the single modality encoder to maintain the size of the model. The results demonstrate the importance of both cross-modal learning and the mixture of two cross-attention functions.

\noindent\textbf{A complementary global visual feature is helpful.} A feature embedding from convolutional neural networks on the whole image can improve our result by around 1\% in document classification. The improvement comes from 1) sometimes document object detector gives an empty detection on low-quality pages, making an all-zero input for our model, so the model learns to exploit global visual feature, and 2) some texts in the proposals are obscured, thus OCR might deliver a random result, making the language embedding uninformative, so an external feature is beneficial.

\begin{table}
    \caption{Fine-tuning with fewer data in document classification.}
    \centering
    \small
    \setlength{\tabcolsep}{5pt}{
    \begin{tabular}{c|c|c}
        \toprule
        
        & w/o VGG-16 & with VGG-16 \\
        
        \midrule
        
        $\text{LayoutLM}_{\text{BASE}}$ (our impl.) & 0.8544 & 0.8712 \\
        SelfDoc & 0.8929 & 0.9150 \\
        
        \bottomrule
    \end{tabular}}
    \label{tab:less}
    \vspace{-2mm}
\end{table}

\noindent\textbf{Fewer data for fine-tuning.} We also provide a scenario where fewer available labels can be accessed in document classification in Table~\ref{tab:less}. To do so, we fine-tune our model and LayoutLM on the validation set of the RVL-CDIP dataset, resulting in an 8 times reduction of fine-tuning data. The VGG-16 model used for fusing is also trained with fewer data. The quantitative results show that our model surpasses LayoutLM by a larger margin compared to fine-tuning with much more data. The observation confirms that our model is also effective when fine-tuning labels are rare.

%-------------------------------------------------------------------------

\section{Conclusion}

We proposed a task-agnostic framework for representation learning and pre-training on document images. Our framework was defined at the semantic components level (rather than words), fully considers the presented property of document data, and includes linguistic, visual, and structural layout information. We employed contextualized learning on the sequential proposals, and encouraged cross-modal learning across language and vision by the proposed cross-modal encoder. We used modality-adaptive attention to emphasize features in language and vision for multimodal fusion. With significantly fewer data for pre-training, we achieved superior performance on multiple tasks.

%-------------------------------------------------------------------------

\section*{Acknowledgments}
This work was supported in part by Adobe Research and NSF OAC 1920147.

%-------------------------------------------------------------------------

{\small
\bibliographystyle{ieee_fullname}
\bibliography{egbib}
}

%-------------------------------------------------------------------------

\appendix

\section{Visualization of Modality-Adaptive Attention}

\begin{figure}[ht]
    \centering
    \includegraphics[width=\columnwidth]{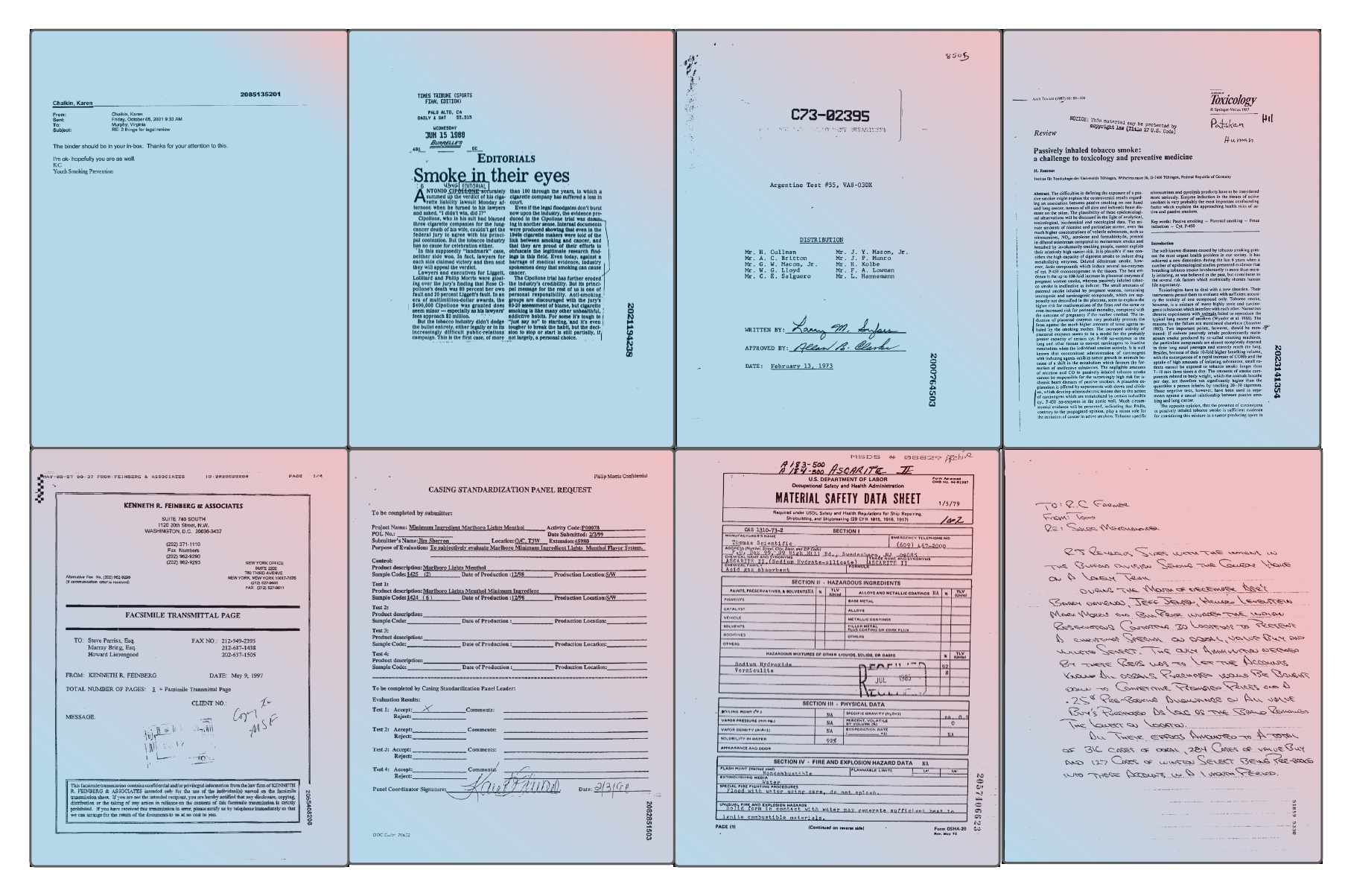}
    \caption{Visualization of Modality-Adaptive Attention on document classification. The size of covering area in blue and red represents the value in $w_\text{lang}$ and $w_\text{visn}$, respectively.}
    \label{fig:maa}
\end{figure}

We visualize the attention mechanism in Fig.~\ref{fig:maa}. The provided samples show that it gives attention scores for each modality adaptively on different types of documents. Some documents with heavy-word or fewer clues in vision have a larger value in $w_\text{lang}$, while some forms with multiple font styles or unrecognizable hand-written enjoy a larger value in $w_\text{visn}$.

\section{Experiments in Document Clustering}

\subsection{Evaluative metrics}

\begin{equation}\nonumber
\begin{split}
& \text{Acc.} = \frac{\sum_{i=1}^n \mathbbm{1}_{{y_i}=\textup{map}(\hat{y_i})}}{n}\ ,\\
& \text{NMI} = \frac{\sum_{i,j} n_{ij}\log \frac{n\cdot n_{ij}}{n_{i+}\cdot n_{+j}}}{\sqrt{(\sum_{i} n_{i+}\log \frac{n_{i+}}{n})(\sum_{j} n_{j+}\log \frac{n_{+j}}{n})}}\ ,\\
\end{split}
\end{equation}
where $\mathbbm{1}$ is the indicator function, and $\textup{map}(\hat{y_i})$ is permutation mapping function that maps each cluster label $\hat{y_i}$ to the ground truth label $y_i$ using linear sum assignment, $n_{ij}$, $n_{i+}$ and $n_{+j}$ represent the co-occurrence number and cluster size of $i$-th and $j$-th clusters in the obtained partition and ground truth, respectively, and $n$ is the total data instance number.
\vspace{2mm}

\subsection{Label sets}

\noindent \# Cluster = 4: \{email, form, handwritten, letter\}

\noindent \# Cluster = 6: \{email, form, handwritten, letter, news article, resume\}

\noindent \# Cluster = 8: \{email, form, handwritten, letter, news article, questionnaire, resume, scientific publication\}

\noindent \# Cluster = 10: \{email, file folder, form, handwritten, letter, news article, questionnaire, resume, scientific publication, specification\}

\noindent \# Cluster = 12: \{email, file folder, form, handwritten, letter, memo, news article, questionnaire, resume, scientific publication, scientific report, specification\}

\end{document}

%% file: math-com.tex
\usepackage{amsmath,amsfonts,bm}

\def\eqref#1{equation~\ref{#1}}
\def\1{\bm{1}}

\def\mH{{\bm{H}}}

\def\mW{{\bm{W}}}

\DeclareMathAlphabet{\mathsfit}{\encodingdefault}{\sfdefault}{m}{sl}
\SetMathAlphabet{\mathsfit}{bold}{\encodingdefault}{\sfdefault}{bx}{n}

%% file: main.bbl
\begin{thebibliography}{10}\itemsep=-1pt

\bibitem{sroie}
Icdar 2019 robust reading challenge on scanned receipts ocr and information
  extraction.
\newblock \url{https://rrc.cvc.uab.es/?ch=13&com=introduction}.

\bibitem{layoutlmgithub2020}
Iit-cdip test collection is unavailable?
\newblock \url{https://github.com/microsoft/unilm/issues/250}, 2020.

\bibitem{arthur2006k}
David Arthur and Sergei Vassilvitskii.
\newblock k-means++: The advantages of careful seeding.
\newblock Technical report, Stanford, 2006.

\bibitem{arxiv}
Arxiv.
\newblock arxiv bulk data access.
\newblock \url{https://arxiv.org/help/bulk_data}, 2020.

\bibitem{ba2016layer}
Jimmy~Lei Ba, Jamie~Ryan Kiros, and Geoffrey~E Hinton.
\newblock Layer normalization.
\newblock {\em arXiv preprint arXiv:1607.06450}, 2016.

\bibitem{beltagy2020longformer}
Iz Beltagy, Matthew~E Peters, and Arman Cohan.
\newblock Longformer: The long-document transformer.
\newblock {\em arXiv preprint arXiv:2004.05150}, 2020.

\bibitem{cosma2020self}
Adrian Cosma, Mihai Ghidoveanu, Michael Panaitescu-Liess, and Marius Popescu.
\newblock Self-supervised representation learning on document images.
\newblock In {\em DAS}, 2020.

\bibitem{dauphinee2019modular}
Tyler Dauphinee, Nikunj Patel, and Mohammad Rashidi.
\newblock Modular multimodal architecture for document classification.
\newblock {\em arXiv preprint arXiv:1912.04376}, 2019.

\bibitem{devlin2019bert}
Jacob Devlin, Ming-Wei Chang, Kenton Lee, and Kristina Toutanova.
\newblock Bert: Pre-training of deep bidirectional transformers for language
  understanding.
\newblock In {\em ACL}, 2019.

\bibitem{girshick2015fast}
Ross Girshick.
\newblock Fast r-cnn.
\newblock In {\em CVPR}, 2015.

\bibitem{gu2020self}
Jiuxiang Gu, Jason Kuen, Shafiq Joty, Jianfei Cai, Vlad Morariu, Handong Zhao,
  and Tong Sun.
\newblock Self-supervised relationship probing.
\newblock In {\em NeurIPS}, 2020.

\bibitem{harley2015icdar}
Adam~W Harley, Alex Ufkes, and Konstantinos~G Derpanis.
\newblock Evaluation of deep convolutional nets for document image
  classification and retrieval.
\newblock In {\em ICDAR}, 2015.

\bibitem{jain2019multimodal}
Rajiv Jain and Curtis Wigington.
\newblock Multimodal document image classification.
\newblock In {\em ICDAR}, 2019.

\bibitem{8892998}
G. {Jaume}, H. {Kemal Ekenel}, and J. {Thiran}.
\newblock Funsd: A dataset for form understanding in noisy scanned documents.
\newblock In {\em ICDARW}, 2019.

\bibitem{katti2018chargrid}
Anoop~R Katti, Christian Reisswig, Cordula Guder, Sebastian Brarda, Steffen
  Bickel, Johannes H{\"o}hne, and Jean~Baptiste Faddoul.
\newblock Chargrid: Towards understanding 2d documents.
\newblock In {\em EMNLP}, 2018.

\bibitem{10.5555/1288165.1288167}
Anthony Kay.
\newblock Tesseract: An open-source optical character recognition engine.
\newblock {\em Linux J.}, 2007(159):2, July 2007.

\bibitem{10.1145/1148170.1148307}
D. Lewis, G. Agam, S. Argamon, O. Frieder, D. Grossman, and J. Heard.
\newblock Building a test collection for complex document information
  processing.
\newblock In {\em SIGIR}, 2006.

\bibitem{liu2019graph}
Xiaojing Liu, Feiyu Gao, Qiong Zhang, and Huasha Zhao.
\newblock Graph convolution for multimodal information extraction from visually
  rich documents.
\newblock In {\em ACL}, 2019.

\bibitem{liu2019roberta}
Yinhan Liu, Myle Ott, Naman Goyal, Jingfei Du, Mandar Joshi, Danqi Chen, Omer
  Levy, Mike Lewis, Luke Zettlemoyer, and Veselin Stoyanov.
\newblock Roberta: A robustly optimized bert pretraining approach.
\newblock {\em CoRR}, abs/1907.11692, 2019.

\bibitem{loshchilov2018decoupled}
Ilya Loshchilov and Frank Hutter.
\newblock Decoupled weight decay regularization.
\newblock In {\em ICLR}, 2018.

\bibitem{lu2019vilbert}
Jiasen Lu, Dhruv Batra, Devi Parikh, and Stefan Lee.
\newblock Vilbert: Pretraining task-agnostic visiolinguistic representations
  for vision-and-language tasks.
\newblock In {\em NeurIPS}, 2019.

\bibitem{marinai2005artificial}
Simone Marinai, Marco Gori, and Giovanni Soda.
\newblock Artificial neural networks for document analysis and recognition.
\newblock {\em TPAMI}, 2005.

\bibitem{mathew2020docvqa}
Minesh Mathew, Dimosthenis Karatzas, R Manmatha, and CV Jawahar.
\newblock Docvqa: A dataset for vqa on document images.
\newblock In {\em WACV}, 2021.

\bibitem{peters2018deep}
Matthew Peters, Mark Neumann, Mohit Iyyer, Matt Gardner, Christopher Clark,
  Kenton Lee, and Luke Zettlemoyer.
\newblock Deep contextualized word representations.
\newblock In {\em NAACL}, 2018.

\bibitem{pramanik2020towards}
Subhojeet Pramanik, Shashank Mujumdar, and Hima Patel.
\newblock Towards a multi-modal, multi-task learning based pre-training
  framework for document representation learning.
\newblock {\em arXiv preprint arXiv:2009.14457}, 2020.

\bibitem{radford2018improving}
Alec Radford, Karthik Narasimhan, Tim Salimans, and Ilya Sutskever.
\newblock Improving language understanding by generative pre-training.
\newblock
  \url{https://s3-us-west-2.amazonaws.com/openai-assets/research-covers/language-unsupervised/language_understanding_paper.pdf},
  2018.

\bibitem{reimers2019sentence}
Nils Reimers and Iryna Gurevych.
\newblock Sentence-bert: Sentence embeddings using siamese bert-networks.
\newblock In {\em EMNLP}, 2019.

\bibitem{ren2015faster}
Shaoqing Ren, Kaiming He, Ross Girshick, and Jian Sun.
\newblock Faster r-cnn: Towards real-time object detection with region proposal
  networks.
\newblock In {\em NeurIPS}, 2015.

\bibitem{shilman2005learning}
Michael Shilman, Percy Liang, and Paul Viola.
\newblock Learning nongenerative grammatical models for document analysis.
\newblock In {\em ICCV}, 2005.

\bibitem{simonyan2014very}
Karen Simonyan and Andrew Zisserman.
\newblock Very deep convolutional networks for large-scale image recognition.
\newblock In {\em ICLR}, 2015.

\bibitem{tan2019lxmert}
Hao Tan and Mohit Bansal.
\newblock Lxmert: Learning cross-modality encoder representations from
  transformers.
\newblock In {\em EMNLP}, 2019.

\bibitem{vaswani2017attention}
Ashish Vaswani, Noam Shazeer, Niki Parmar, Jakob Uszkoreit, Llion Jones,
  Aidan~N Gomez, {\L}ukasz Kaiser, and Illia Polosukhin.
\newblock Attention is all you need.
\newblock In {\em NeurIPS}, 2017.

\bibitem{wei2013evaluation}
Hao Wei, Micheal Baechler, Fouad Slimane, and Rolf Ingold.
\newblock Evaluation of svm, mlp and gmm classifiers for layout analysis of
  historical documents.
\newblock In {\em ICDAR}, 2013.

\bibitem{wu2019detectron2}
Yuxin Wu, Alexander Kirillov, Francisco Massa, Wan-Yen Lo, and Ross Girshick.
\newblock Detectron2.
\newblock \url{https://github.com/facebookresearch/detectron2}, 2019.

\bibitem{xie2017aggregated}
Saining Xie, Ross Girshick, Piotr Doll{\'a}r, Zhuowen Tu, and Kaiming He.
\newblock Aggregated residual transformations for deep neural networks.
\newblock In {\em CVPR}, 2017.

\bibitem{xu2020layoutlm}
Yiheng Xu, Minghao Li, Lei Cui, Shaohan Huang, Furu Wei, and Ming Zhou.
\newblock Layoutlm: Pre-training of text and layout for document image
  understanding.
\newblock In {\em SIGKDD}, 2020.

\bibitem{yang2017learning}
Xiao Yang, Ersin Yumer, Paul Asente, Mike Kraley, Daniel Kifer, and C
  Lee~Giles.
\newblock Learning to extract semantic structure from documents using
  multimodal fully convolutional neural networks.
\newblock In {\em CVPR}, 2017.

\bibitem{zhong2019publaynet}
Xu Zhong, Jianbin Tang, and Antonio~Jimeno Yepes.
\newblock Publaynet: largest dataset ever for document layout analysis.
\newblock In {\em ICDAR}, 2019.

\end{thebibliography}
